\documentclass[conference]{IEEEtran}
\IEEEoverridecommandlockouts
\usepackage{cite}
\usepackage{amsmath,amssymb,amsfonts}
\usepackage{algorithmic}
\usepackage{graphicx}
\usepackage{textcomp}
\usepackage{xcolor}
\def\BibTeX{{\rm B\kern-.05em{\sc i\kern-.025em b}\kern-.08em
    T\kern-.1667em\lower.7ex\hbox{E}\kern-.125emX}}
\usepackage[hidelinks]{hyperref}
\usepackage{orcidlink}

\begin{document}

\title {Analyzing the Impact of Adversarial Examples on Explainable Machine Learning\\
}

\author{\IEEEauthorblockN{Prathyusha Devabhakthini$^{\lor}$, Suvendu Nayak$^{\land}$, Raj Shukla$^{\lor}$,  Sasmita Parida$^{\land}$, and Tapadhir Das*}
\IEEEauthorblockA{$^{\lor}$School of Computing and Information Science, Anglia Ruskin University, Cambridge, UK
\\
$^{\land}$Department of Computer Science and Engineering, Silicon University, Odisha, India
\\
*Department of Computer Science, University of the Pacific, Stockton, USA
}
\IEEEauthorblockA{
Email: \{pd483@student,raj.shukla\}@aru.ac.uk, \{suvendu2006,sasmitamohanty\}@gmail.com, tdas@pacific.edu
}}

\maketitle

\begin{abstract}
Adversarial attacks are a type of attack on machine learning models where an attacker deliberately modifies inputs to cause the model to make incorrect predictions. Adversarial attacks can have serious consequences, particularly in applications such as autonomous vehicles, medical diagnosis, and security systems. Work on the vulnerability of deep learning models to adversarial attacks has shown that it is very easy to make samples that make a model predict things that it doesn’t want to. In this work, we suggested a model-agnostic explanation-based method to find adversarial samples in two datasets: Rotten Tomatoes and IMDB movie reviews. We used Local Interpretable Model- Agnostic Explanations (LIME) explainability techniques to look at attacks made with TextAttack. We observed that on the rotten tomato datasets, 77 \% of the time, an adversarial attack was detected 77 \%. Similarly, on the IMDB dataset, we got an accuracy rate of 88 \% for adversarial attacks through simulation. The proposed model provides better outcomes for identifying adversarial attacks.
\end{abstract}

\begin{IEEEkeywords}
Adversarial Attack, Natural Language Processing, Text Classification, Explainability, Local Interpretations
\end{IEEEkeywords}

\section{Introduction}
Artificial intelligence (AI) and natural language processing (NLP) have come a long way in recent years, but there are still concerns regarding the ramifications of using them to automatically regulate data in many situations. One concern is that automated control of material on social media networks will accelerate the development of AI, which can offset the impacts of AI. Let that be an image, speech, or text that the models are trained on; these can be attacked. For example, a small fraction of noise added to an image trained by a model can completely mispredict the intended result. This can impact many applications such as self-driving cars, voice assistants, chatbots, and many other applications. Additionally, these models are facing a new challenge: adversarial attacks that are designed to be imperceptible to humans, while still causing the model to misclassify the input.  The discovery of adversarial examples in neural networks has led to tremendous growth in the field \cite {b1}.

In recent years, there has been increased interest in designing adversarial attacks for NLP tasks to create attack vectors and produce corresponding solutions to make the model robust. Learning the impacts of attacks with explainability is a solution to measuring and learning the effects on AI models. Adversarial examples of text data, a technique for producing adversarial perturbations on text classification, and their impact on model explainability are presented in \cite{b2}. The adversarial samples in this research are purposely generated sentences with English transformations, such as misspellings, replacing them with synonyms, which can lead the classifier to provide incorrect prediction outputs, as shown in Table~\ref{tab: sample adversarial examples}. When the above sentence is changed semantically, while the meaning is the same, it does not make any difference for a person to comprehend. However, the trained model will classify the sentence with 100\% confidence as belonging to a different class.

\begin{table}
\centering
\setlength{\tabcolsep}{0.5em}
\renewcommand{\arraystretch}{2}
\caption{Adversarial example on text data \cite{b2}}
\begin{tabular}{|p{4.8em}| p{11.9em}|p{8.2em}|}
 \hline
 \textbf{Original Input} & [[\textbf{offers}]] that [[\textbf{rare}]] combination of entertainment and education. & [[Positive(100\%)]] \\
 \hline
 \textbf{Adversarial Example} & [[\textbf{prescribes}]] that[[\textbf{sparse}]] combination of entertainment and education. & [[Negative(100\%)]] \\ [1ex]
 \hline
 \end{tabular}
\label{tab: sample adversarial examples}
\end{table}

A theoretical model of the adversarial example-crafting process is challenging to construct, which makes it tough to defend against it. As there are currently no accepted theoretical frameworks for understanding the solutions to these complicated performance issues, building any prospective defensive solution is exceedingly difficult. Additionally, defending against adversarial samples is tough since machine learning (ML) models are required to produce good results for every potential input. The majority of the time, ML models function well, yet they only operate on a relatively tiny subset of all possible inputs. If one can successfully block one kind of attack, it leaves another vulnerability exposed to the attacker who is aware of the attack being utilized. Making a defense that is capable of protecting against a strong and adaptable attacker is an important research area \cite{b3}. 

In addition to initiating attacks, recent works have studied the effects of model explainability before and after attacks. Recently, there has been a significant increase in the development of explainable machine learning techniques. This is mostly due to increased awareness among regulators and end-users of the impact of machine learning models and the necessity to comprehend their conclusions. The objective is for the user to comprehend the predictions of ML models, which is accomplished through explanations. Explainability is a technique for analyzing any ML model that involves perturbing the input of data samples and monitoring how forecasts change over time \cite{b4}. The output is a list of explanations, which interpret how each feature contributes to data prediction, highlighting which adjustments to the features will have the greatest influence on the forecast \cite{b5}. 

The semantic similarity model, like the model under attack, is subject to adversarial examples. It is, thus, difficult to determine which model committed the mistake when a legitimate adversarial example is discovered, as it was either the model being attacked or the model employed to impose the constraint. Storing the attacks and predicting the results for models can help distinguish attacks and model errors. The explainability of the model helps understand when an attack is generated and what features make the model classify incorrectly. In this way, future models can be developed to easily explain and understand the attacks. The other limitation of current research is that while current efforts have investigated adversarial attacks and explainability in ML models, there is limited research combining them with NLP \cite{b6}. 

In this paper, we have examined the impact of adversarial examples on the explainability of the ML model within NLP. Our work develops a safe DNN model and analyzes adversarial situations and their effects, providing a solution to help better understand DNNs, thereby improving their reliability and safety. The use of adversarial instances provides a new domain in which to investigate the attractive aspects of DNNs. Additionally, our proposed method combines two concepts: adversarial attacks and explainability, to explain the model's performance after the attacks. Our key contributions are listed below: 
\begin{itemize}
\item We propose to explain adversarial attacks with explainability to build a robust model against adversarial attacks on text data.  
\item We experiment with adversarial attacks and then compare the results. We train a binary text classification model on two different datasets and use pre-trained transformer models for comparison.
\item We generate adversarial examples on the model and the pre-trained model. We explained the original model and attacked models with explainable AI techniques for analyzing the effect of attacks on text sentences
\item We analyze the effect of the attack on explainability on three state-of-the-art transformer-attacked models, BERT, RoBERTa, and xlnet-based models.
\end{itemize}
Moreover, this work aims to give answers to the following research questions: 
\begin{itemize}
    \item How is the explainability of the model affected by adversarial attacks?
    \item How are the success rates of attacks on each trained model?
    \item What features are mostly compromised by the adversarial attacks?
\end{itemize}

The rest of this work is arranged in the following manner:  Section ~\ref{literature} addresses the literature as well as earlier comparable works. Section ~\ref{Background} provides a brief background on the different concepts used in this research. Subsequently, Section ~\ref{methodology} presents the proposed integrated pipeline impacted by the adversarial attack and incorporating XAI techniques. Section ~\ref{results} provides detailed results and a discussion of our findings. Finally, section ~\ref{conclusions} gives a summary of our conclusions as well as suggestions for future research directions.

\section{Literature Review} \label{literature}

In this section, we present recent work, more specifically on adversarial attacks, adversarial attacks on text data, and explainable ML models by researchers. Adversarial attacks create distorted copies of the input data that deceive the classifier (i.e., modify its output). These attacks can be applied to any data. For example, producing noise in image data is an adversarial attack on image data.

\subsection{Adversarial Attacks}
Adversarial attacks on different data are detailed in a study by Xi et al. in \cite{b8}, where the authors generated adversarial assaults using three distinct data types (images, graphs, and text) and then explained the solutions. Carlini et al. used adversarial attacks on speech data to successfully convert any audio waveform into any target transcription with only little distortion \cite{b9}. Xie et al. discussed that the adversarial attack on the image recognition model might cause it to produce an explanation that emphasizes bigger models. The authors proposed AdvProp, which is an enhanced method based on auxiliary batch normalization to avoid overfitting problems. It divided the bigger model into mini-batches and then training was applied \cite{b10}. Slack et al. presented a scaffolding scheme that hides the effect of biases on the classifier with the adversarial object \cite{b11}. Real-world data was considered, and the post hoc method was applied to the adversarial classifier to avoid biases. Here, three datasets, such as COMPAS, Communities and Crime, and German Credit, were trained and evaluated. This method claimed exploitation for a biased classifier using a perturbation method where the input was biased and the final classifier was controlled behavior. 

Aryal et al. presented a study on adversarial attacks in malware analysis that involve modifying the malware sample in such a way that it can bypass detection by antivirus software  \cite{b12}. It could be by modifying the binary code of the malware, inserting additional code to confuse the antivirus software, or modifying the metadata of the file to make it appear legitimate. Adversarial perturbation was a solution by adding small, carefully crafted modifications to the malware sample that are designed to cause the antivirus software to misclassify the sample. Sungtae et al. proposed LAVA adversarial attacks and discussed the performance of attacks on longitudinal data from the Electronic Health Record (EHR) using various models, claiming that LAVA adversarial attacks are more difficult to detect than other attacks \cite{b13}. LAVA avoids selecting features that are highly relevant to the prediction goals by default. Open-source packages for generating adversarial attacks on NLP models are described in \cite{b14}, \cite{b15}, \cite{b16}. 

Miyato et al proposed one of the first studies to use adversarial training for natural language processing problems, executing perturbations at the word embedding level rather than at the level of actual input space \cite{b17}.  Attack word embedding is a technique for adding perturbation to the word embedding to deceive a classification algorithm. Similarly, when it comes to phrases, words, and letters, changing a single letter in a sentence changes the model's predicted outcome. The attack technique can do this by identifying the most impactful letter substitution using gradient information. Xu et al. \cite{b8} proposed TextTricker based on a loss and gradient white box adversarial attack model used to produce incorrect explanations on a text classification model, which might cause it to produce an explanation that emphasizes irrelevant words, leading the user to make incorrect assumptions. It supported targeted as well as non-targeted attacks and evaluated two datasets with a noticeable success rate. The work done by Samanta et al. marks the beginning of the process of creating adversarial statements that are grammatically accurate and retain the syntactic structure of the original phrase, respectively \cite{b18}. The authors accomplish this by substituting synonyms for original words or by including certain words with varying meanings in various situations. Some of the toolkits for producing attacks on text data for generating attacks on NLP are OpenAttack and TextAttack \cite{b19}, \cite{b2}.

\subsection{Explainable Machine Learning Models}
The method or algorithm that creates the explanations is a model for interpretability. Much research is being conducted to develop new methodologies and tactics to improve interpretability while limiting the reduction in projected accuracy.  Ribeiro et al. introduce the Local interpretable Model-Agnostic Explanations (LIME) model and explain how it delivers many of the desired properties for interpretability in \cite{b20}. They also discuss some of the major difficulties in explaining data, as well as a newly announced model-agnostic explanation technique (LIME) that overcomes these difficulties. The importance, evaluation, and properties of interpretability with models and metrics are explained.  The interpretability models for both white-box and black-box models are explained with their functionalities toward different types of data, how the models have been developed, offering explanations, comparisons, and mainly focusing on the literature for various explanations used in different papers. Linardatos et al. stated that the LIME and Shapley Additive Explanations (SHAP) approaches are the most comprehensive and widely used methods for visualizing feature interactions and significance in the literature \cite{b21}. Rosenberg et al. focused on the concept of adversarial attack using the transferability of explainability. In the first phase, it analyses the features of the malware classifier using explainability, and then an adversarial attack in a black-box scenario was added and trained \cite{b22}. The authors claimed the selection of features to be modified in the presence of adversarial examples using explainability for malware classifiers.

\subsection{Adversarial attacks on interpretability}
Watson et al. proposed research on explainability-based machine learning models using SHAP on Electronic Health Records (EHR) and Chest X-Ray (CXR) data \cite{b23}. They were able to characterize how various attack tactics operate on various datasets and use this knowledge to identify samples that have been manipulated in an adversarial manner using SHAP values. The efficacy of attacks is looked into six models, and the results of adversarial sample detection are evaluated, with strategies compared against advanced MLLOO explainability-based adversarial detection methods. From the literature, it is evident that there are tools to generate adversarial attacks on text data. Along with that, there are methods to improve the interpretability of the ML models, such as LIME and SHAP. The combination of these two concepts in Natural Language Processing is useful for building better complex models and improving understanding of Adversarial attacks on text data. In addition, the explanations make users choose better model-based explanations. 

\section{Background} \label{Background}

\subsection{Adversarial Attacks} \label{TextAttack}
In the past, researchers have explored adversarial training for NLP models in a variety of ways, with various objective functions, restrictions, search techniques, transformations, and search algorithms being used to produce adversarial instances ~\cite{b24}.

\subsubsection{White-box Attacks} \label{White-box Attack}
In this type of attack model, total information is used for classification. The attacker has knowledge about the training method using optimization and training distribution with minimization of error rate. 

\subsubsection{Black-box Attacks} \label{Black-box Attack}
Here, it is assumed that no information about the model is available to the attacker. The adversary method exploits the crafted input along with the outputs. Further subdivided into the following types:

\subsubsection{Non-adaptive Black-box Attacks} \label{Non-adaptive Black-box Attack}
This model can only use training data distribution and an algorithm that operates on local input and is targeted to output. 

\subsubsection{Adaptive Black-box Attacks} \label{Adaptive Black-box Attack}
In this model, an adaptive adversary method is used without any complete knowledge of the training method. Generally, it's similar to a plaintext attack in the case of cryptography.

\subsubsection{Strict Black-box Attacks} \label{Strict Black-box Attack}
In this adversary method, training data does not follow the distribution; rather, it processes data in pairs and produces targeted output.

\subsection{Explainability} \label{Explainability}
Explainable ML approaches have recently grown in popularity. This is due to regulators and end users being more aware of the influence of ML models and the need to understand their findings. Explanations help the user comprehend the ML models' predictions. The explanations are models for interpretability. Much research is being done to promote interpretability in prediction tasks while limiting the reduction in projected accuracy. This is accomplished by providing explanations to help the user comprehend the predictions of machine learning models. To do this, we employ the explainability technique, which is an algorithm that creates possible explanations.

According to the literature, interpretability is comprised of three aims, all of which are interconnected and often compete with one another:

\begin{itemize}
\item Accuracy: Rather than a theoretical link, accuracy refers to the actual connection between the explanatory strategy and the ML model prediction. If this goal is not achieved, the explanation will be rendered ineffective since it will not be loyal to the prediction that it is intended to explain.

\item Understandability: It is connected to the ease with which an explanation is grasped by the spectator and is measured in terms of observers. This is an important aim since, no matter how precise an explanation is, it is worthless if it is not easily comprehended.

\item Efficiency: It refers to the amount of time it takes for a user to comprehend an explanation. If this requirement is not met, it is possible to argue that, given an endless amount of time, practically any model is interpretable. As a result, an explanation should be intelligible in a limited length of time, ideally less than a few minutes. This aim is connected to the preceding one, understandability: in general, the more intelligible an explanation is, the more swiftly it is comprehended.
\end{itemize}

\subsubsection{Model agnostic explanation methods}
A model-agnostic explanation strategy is a method of explaining that is not dependent on a particular model. Even though these methods are more useful when applied to a specific model and case, explanation methods have a distinct advantage in that they are completely separate from the original model class, allowing them to be applied to completely different use cases even when the predictive model is the same. The goal is to make the user understand the predictions of ML models, which is achieved through explanations. For this, we make use of an explanation method, which is nothing more than an algorithm that generates explanations.  Several approaches are known as LIME explanations, which stands for Local Interpretable Model-agnostic Explanations \cite{b20}. 

\section{Methodology} \label{methodology}
\begin{figure*}[t]
    \centering
    \includegraphics[width=0.7\linewidth]{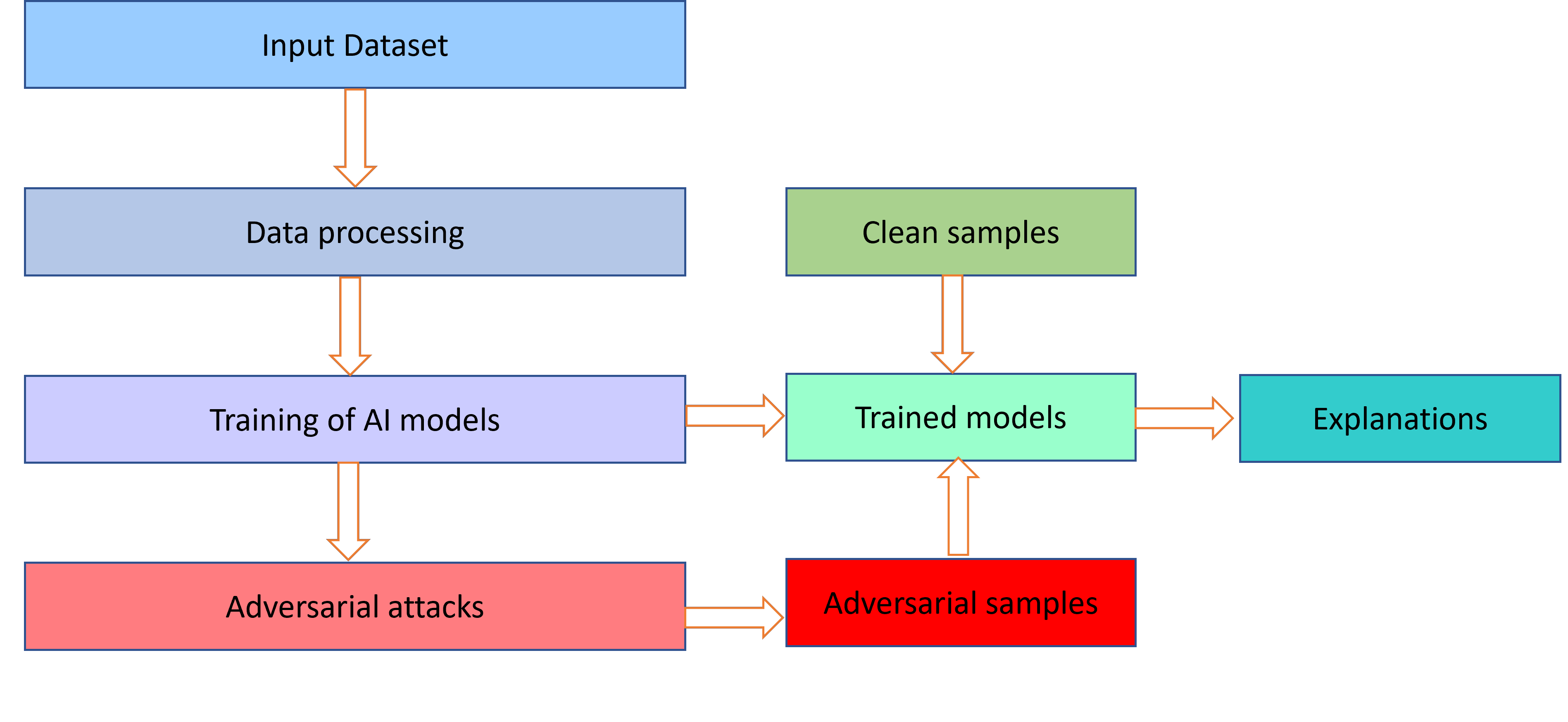}
    \caption{Workflow diagram of the proposed model}
    \label{fig:workflow_diagram}
\end{figure*}

Figure ~\ref{fig:workflow_diagram} presents the schematic diagram of the proposed framework.  As illustrated in the figure, in the initial stage of our proposed method, we prepare the model for training. This involves conducting initial data analysis and performing data cleaning to ensure the data is suitable for training. We perform the following steps for data preparation and cleaning.

\subsection{Input Dataset}
For the intended dataset, any labeled NLP dataset can be utilized for this work. 

\subsection{Data Preprocessing}
\subsubsection{Noise Removal} This is the initial stage that involves removing all formatting from text, like HTML tags, paragraph or page breaks, as well as punctuation marks.

\subsubsection{Tokenization}
Tokenization is the act of identifying the fundamental elements in a linguistic expression that do not need to be deconstructed. In NLP, tokens are words that are separated by spaces, and the act of breaking down sentences into tokens, which may be words or punctuation marks.

\subsubsection{Normalization}
Text may include a variety of non-standard token types, including digit sequences, acronyms, words, as well as letter sequences in all capital letters, abbreviations, roman numerals, mixed-case words, URLs, and email addresses. The process of normalization is the rewriting of such material in everyday language. In NLP, stemming \& lemmatization are generally used for text normalization. Stemming is the process of removing prefixes and suffixes from words, whereas lemmatization is the process of reducing a word to its most fundamental form. 

\subsubsection{Stop Word Removal}
Stop words are terms in language that are regularly found in numerous sentences and are therefore taken out either before or after the natural language data processing process is completed.  Even though stop words are frequently encountered in NLP tasks such as sentiment analysis and topic modeling, their frequency may result in model bias against the space containing these words, resulting in a bad model for these tasks. To address this issue, such phrases are removed before the analysis of natural language.

\subsubsection{Word Embeddings/Vectorization}
The process of converting a text into a vector is called word embedding/vectorization. Some of the methods used in NLP for this are: i) TF-IDF (Term Frequency – Inverse Document Frequency), ii) n-gram-based approach, and iii)  GloVe. 

\subsection{ML Model Training}
After data pre-processing, the model is then trained using AI algorithms. Our focus in this step is to create a reliable model that can be used as input for generating attacks and explanations. Alongside model training, we also evaluate its performance and generate explanations for a subset of examples from the test dataset. These generated explanation examples will later be compared to the explanations of the same examples from the attacked model in the final step.

\subsection{Adversarial Attacks}
After training, we generate adversarial attacks. Here we perturb the text that tricks the NLP model into making incorrect predictions while adhering to specific constraints. To generate attacks, we follow a series of steps, including selecting the model, wrapping the model using the model wrapper, choosing the dataset, creating an attack, and finally attacking the model. We explore different attack models to compare their performance in generating attacks.  For generating adversarial samples, we utilized the textAttack tool as it provides various recipes, namely TextFooler and TextBugger \cite{b2}.

\subsection{Explainability}
In the final step, we generate explanations for the attacked model and analyze the effects of attacks on input sentences. We have used the LIME (Local Interpretable Model-Agnostic Explanations) method for generating explanations \cite{b20}. LIME operates on the principle of zooming into the local region of each prediction, allowing us to obtain valid explanations without needing to consider the entire model. By fitting a linear interpretable model, known as a surrogate, in the local area of interest, LIME provides a local approximation of the complex model. LIME solely relies on the inputs and outputs of the model to generate explanations. This approach allows us to gain insights into the interpretability and understanding of the attacked model's predictions. By comparing the explanations of the attacked model with those of the original model, we can identify any changes or vulnerabilities introduced through adversarial attacks. Through the usage of explanations, we can analyze how the attacks influence the behavior of the attacked model, thereby assessing its robustness and reliability. This step provides valuable insights into the differences between the original and attacked models, highlighting any modifications in their decision-making processes. By analyzing the effects of the attacks on the model's explanations, we can better understand the implications of adversarial inputs and improve our understanding of the model's behavior in real-world scenarios.

\section{Results and Discussions} \label{results}
In this section, we explain the detailed results of the various steps. We begin by describing the data set used in this study. 

\subsection{Data set}
For this research, we utilized two datasets: 1) Rotten Tomatoes Movie Reviews (MR) and 2) the IMDB dataset.  The first dataset, Rotten Tomatoes, was collected by Pang and Lee \cite{b25}. It comprises 5,331 movie reviews, with positive and negative sentiments. Each review consists of sentences with an average sentence length of 32 words. The second dataset, IMDB, was collected by Maas et al \cite{b26}. It encompasses 50,000 positive and negative reviews, with an average sentence length of 215.63 words.

\subsection{Neural Network Design}
We employ state-of-the-art pre-trained transformer models. Since our approach focuses on adversarial training concerning model explainability, we choose the best-performing pre-trained model for text classification from Hugging Face Transformers' pre-trained models. We have used BERT, RoBERT, and XLNet-based cased models and have compared the attacks with success rates on both datasets. The pre-trained models from Transformers can be loaded along with the tokenizer to generate attacks and for explanations. The accuracy performance of the models is shown in the table~\ref{table:accuracy_comparison}. 

\begin{table}[t]
\centering
\setlength{\tabcolsep}{0.5em}
\renewcommand{\arraystretch}{2}
\caption{Table comparing accuracy of pre-trained models}
\begin{tabular}{|p{1.4cm}| p{2.5cm}|p{1.4cm}|}
 \hline
 \textbf{Model} & \textbf{Rotten Tomatoes} & \textbf{IMDB} \\ [0.8ex]
 \hline 
 BERT & 87.60\% & 91.90\% \\
 \hline 
 RoBERTa & 89.90\% &  94.10\% \\
 \hline
 XLNet & 90.80\% & 95.70\% \\ [1ex]
 \hline
 
\end{tabular}
\label{table:accuracy_comparison}
\end{table}

\subsection{Generating attacks for pre-trained Models}
We have used transformer models to compare and evaluate the effect of an attack on a single sentence for visualization of the explanation results. As a result, the attack results are displayed with 10 sample sentences from the dataset. The results of each model are different in each sentence because of the difference in components used for producing the attacks. For example, in the results explained with the RoBERTa model, the attacks are performed on 10 samples from the dataset. Out of 10, the initially trained model has 1 wrong prediction out of 10 with model accuracy 90\%. After the attack, there were 0 failed attacks leading to a success rate of 100\%.  Among those 9 successful attacks performed by the attack recipe, on average, 15.2\% of words are changed to change the prediction results, and made 113.89 queries to find successful perturbation. The average number of words in 10 samples is 19.57 per sentence. The Original Accuracy for BERT-based-uncased and Xlnet-based-cased models is 100\% for those ten samples. Since the model is the least performing, the number of words changing for producing the attacks is also less for SVM. 

\subsection{Explainations with LIME}
Explanations with LIME on text data provide probabilities for each text and are visualized with different colors, with a score for each feature in the sentence. The higher the score with the respective class color, the more it affects the prediction rate in a sentence. Since it is a binary classification to predict positive or negative reviews from datasets. The class names are categorized into 0 and 1.  For a sample text, the probabilities for each word are calculated, and then the class names are explained as input parameters.

\begin{figure*}[t]
\begin{center}
\includegraphics[width=0.9\linewidth]{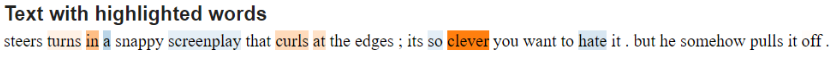}
\caption{Sample explanation }
\label{sample_explanation}
\end{center}
\end{figure*}

\begin{figure}[t]
\begin{center}
\includegraphics[width=0.4\textwidth]{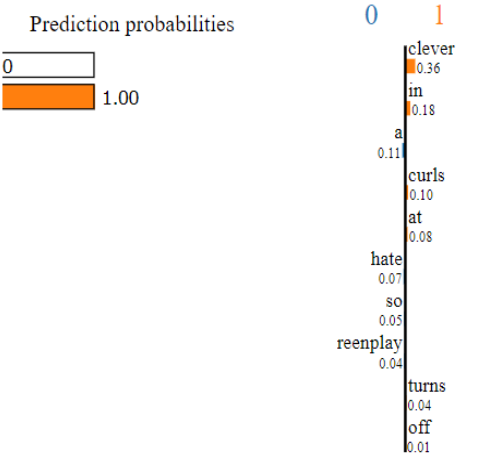}
\caption{Sample feature explanation }
\label{sample_explanation_feature}
\end{center}
\end{figure}

The number of features in the text is considered to be ten, as the average number of words in the ten samples is 19.5 words. The sample text explanations are represented as shown in the figures~\ref{sample_explanation} and \ref{sample_explanation_feature}. In figure~\ref{sample_explanation_feature}, the orange highlights in \ref{sample_explanation} are for class 0 and blue are for class 1. The darker the highlight is, the more the feature importance for that class. The feature importance values are shown in the figure below for the whole text. For example, the word "\textbf{clever}'' highlighted with 36\% feature importance is making the model classify the input sentence as a positive movie review. It shows which word is playing an important role in predicting the output. This is the main reason to analyze the importance of explanations for analyzing the attacks as well. One can see a change in words, and the importance can affect the outputs. 

\subsection{Analyzing the explanations with different models}
The explanations for each model are different before and after the attack because the way the model interprets words is different. The main aim of this section is to explain how the different models are affecting the model explanations before and after the attack. We have used similar explanations for different ML  models - BERT-based-uncased, RoBERTa, and XLNet-based-cased models. To analyze the difference in explanations, we selected a single sample input sentence: 

\textbf{Sentence}: \textit{“steers turns in a snappy screenplay that curls at the edges; it's so clever you want to hate it. But he somehow pulls it off .”}

\textbf{Label}: 1 (Positive Review)

\begin{table}[t]
\centering
\setlength{\tabcolsep}{0.5em}
\renewcommand{\arraystretch}{2}
\caption{Comparing three models based on single sample text prediction}
\begin{tabular}{|p{2cm}| p{2.7cm}|p{2.7cm}|}
 \hline
 \textbf{Original Model} & \textbf{Confidense score for class 0} & \textbf{Predicted Probability for class 1} \\ [0.8ex]
 \hline
 BERT-based-uncased & 0.00 & 1.00 \\
 \hline 
 RoBERTa & 0.06 &  0.94 \\
 \hline
 XLNet-based-cased & 0.09 & 0.91 \\ [1ex]
 \hline
\end{tabular}
\label{tab: comparison_before_attack}
\end{table}

The confidence scores of each model on this sentence are explained in the Table ~\ref{tab: comparison_before_attack}. Although these models predicted the review as positive with more probability, there are huge differences in word importance in each model. After the attack, all 3 models predicted the same sentence as negative but with different probabilities as shown in the Figure ~\ref{tab:comparison_after_attack}. The Table ~\ref{tab:comparison_change} shows the change in probabilities of each class for models as shown in the table. 

Figure 5-7 shows the feature importance given to sentences by each model before and after the attack. It is observed that \textbf{"clever'', "pulls''} are given importance for 3 models to classify the sentence as positive. And \textbf{"hate'', "but''} are given more importance to classify them as negative. It is important to analyze what words are given more priority in these explanations. As all the explanations visualized are for the original model, these explanations change when the model is attacked. After the attack, the explanations and word importance are changed because when the model is attacked, the words in the original sentence are replaced with transformations like synonyms, changes in spelling, or by removing suffixes/prefixes.

\begin{table}[t]
\centering
\setlength{\tabcolsep}{0.5em}
\renewcommand{\arraystretch}{2}
\caption{Comparing three models based on single sample text prediction after attacking the sentence
}
\begin{tabular}{|p{2cm}| p{2.7cm}|p{2.7cm}|}
 \hline
 \textbf{Attacked Model} & \textbf{Confidence score for class 0} & \textbf{Predicted Probability for class 1} \\ [0.8ex]
 \hline
 BERT-based-uncased & 0.96 & 0.04 \\
 \hline
 RoBERTa & 0.61 &  0.39 \\
 \hline
 XLNet-based-cased & 0.68 & 0.32 \\ [1ex]
 \hline
 
\end{tabular}
\label{tab:comparison_after_attack}
\end{table}

It is clear from the results that the BERT model is more affected by perturbations than the other two models. The other two models were affected nearly the same (60\%).  The example positive review with 100\% confidence is converted to a 93\% confident negative review after an attack on the BERT-based-uncased model. Similarly, the table~\ref{tab:comparison_after_attack} explains how the confidence values for the example sentence are changed from positive class to negative class.

\begin{table}[t]
\centering
\caption{ Comparison of change in values after attack
}
\setlength{\tabcolsep}{1em}
\renewcommand{\arraystretch}{2}
\begin{tabular}{|p{1.8cm}| p{5cm}|}
 \hline
 \textbf{Model} & \textbf{Confidence Values of each class}  \\ [0.8ex]
 \hline
 BERT-based-uncased & [[Positive (100\%)]] to [[Negative (93\%)]] \\
 \hline 
 RoBERTa & [[Positive (94\%)]] to [[Negative (66\%)]]\\
 \hline
 XLNet-based-cased & [[Positive (85\%)]] to [[Negative (89\%)]]\\ [1ex]
 \hline
 
\end{tabular}
\label{tab:comparison_change}
\end{table}

The most affected BERT-based-uncased model completely changes the prediction with 93\% confidence as a negative class.  It can be analyzed from the results that the BERT model on the MR dataset predicts the wrong output with 93\% confidence, stating it as the most affected one. 

\begin{figure}[t]
\begin{center}
\includegraphics[width=0.45\textwidth]{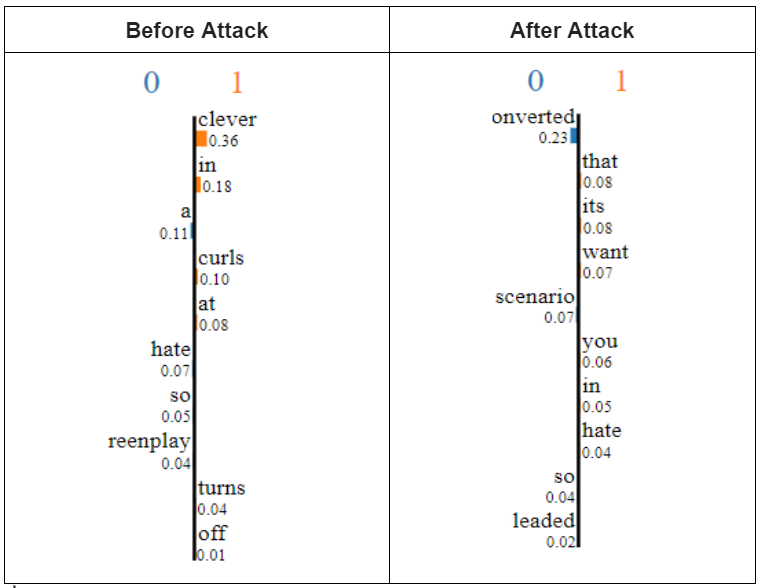}
\end{center}
\caption{This figure illustrates the word probabilities in a sample sentence before and after the attack for the BERT model.}
\end{figure}

\begin{figure}[t]
\begin{center}
\includegraphics[width=0.45\textwidth]{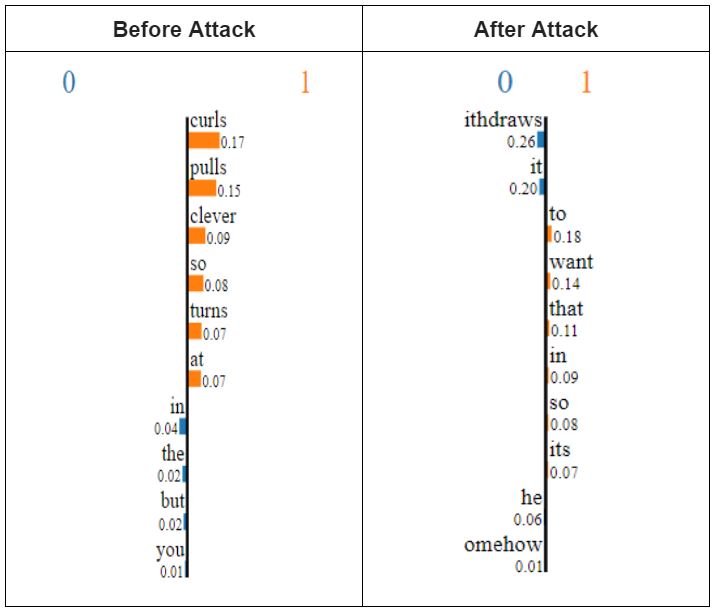}
\end{center}
\caption{This figure illustrates the word probabilities in a sample sentence before and after the attack for the RoBERTa model.}
\end{figure}

\begin{figure}[t]
\begin{center}
\includegraphics[width=0.45\textwidth]{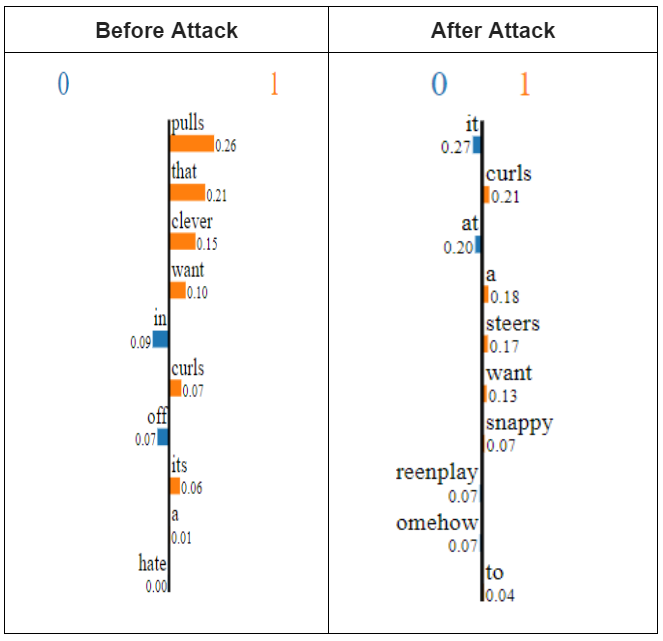}
\end{center}
\caption{This figure illustrates the word probabilities in a sample sentence before and after the attack for the XLNet model.}
\end{figure}

It can be analyzed that on small datasets, the attacks with RoBERTa are predicting wrong with less probability when compared to the BERT model. The effect of the BERT model is greater than that of the other two models. When comparing both datasets, the RoBERTa model was less attacked in the sample example sentence compared to other models in both datasets, as shown in the table~\ref{tab:comparison_dataset}. 

\begin{table}[t]
\centering
\caption{ Comparison of IMDB and MR with Roberta
}
\setlength{\tabcolsep}{0.8em}
\renewcommand{\arraystretch}{2}
\begin{tabular}{|p{1.7cm}| p{2.4cm}|p{2.4cm}|}
 \hline
  & \textbf{IMDB dataset} & \textbf{MR dataset}  \\ [0.8ex]
 \hline
 \textbf{Before Attack} & 6\% for negative, 94\% for positive & 6\% for negative, 94\% for positive\\
 \hline 
 \textbf{After Attack} & 61\% for negative, 39\% for positive & 61\% for negative, 39\% for positive\\
 \hline

\end{tabular}
\label{tab:comparison_dataset}
\end{table}

\section{Conclusions} \label{conclusions}
In conclusion, our model has analyzed the effects of explainability of ML models with Local Interpretable Model-Agnostic Explanations on adversarial examples generated on text data. It is evident that with the help of explanations, any black-box model predictions can be explained to humans. In real-world NLP, the possibility of adversarial examples in the data is high, and the need to create solutions is really important. This not only helps humans understand the AI model but also increases their trust in the output from AI models, and highlights the usage of explainable ML models along with the development of ML models to build robust models. This explainability concept can be included in any project to generate explanations for each prediction. Our model can avoid mistakes from automated ML models and can be further implemented to alert users about changes in word importance and direct users about adversaries. In our work, we consider that most
of the real-time NLP datasets contain 5,000+ samples with a variety of features. This implementation allows users to understand the concept of adversarial attacks and to understand the usage of explainability algorithms to analyze the effect of adversarial attacks on text data. We used different ML algorithms and transfer learning-based models to analyze the impact of adversarial attacks on model explainability. For future work, the presented system can be further developed into a more sophisticated system, such as automated recognition of adversarial attacks. Our proposed model could be examined with larger datasets and analysed with more features.

\end{document}